\documentclass[conference,transmag]{IEEEtran}
\IEEEoverridecommandlockouts
\usepackage{cite}
\usepackage{amsmath,amssymb,amsfonts}
\usepackage{algorithmic}
\usepackage{graphicx}
\usepackage{textcomp}
\usepackage{booktabs}
\usepackage{xcolor}
\usepackage{array}
\usepackage{multirow}
\usepackage{amssymb}
\usepackage{marvosym}
\usepackage{ifsym}
\usepackage[hidelinks]{hyperref}       
\hypersetup{
	colorlinks=true,
	linkcolor=blue,
	filecolor=red,      
	urlcolor=red,
	citecolor=green,
}
\usepackage[T1]{fontenc}
\def\BibTeX{{\rm B\kern-.05em{\sc i\kern-.025em b}\kern-.08em
    T\kern-.1667em\lower.7ex\hbox{E}\kern-.125emX}}
\begin{document}

\title{TP-UNet: Temporal Prompt Guided UNet for Medical Image Segmentation}

\author{\IEEEauthorblockN{Ranmin Wang\IEEEauthorrefmark{1},
Limin Zhuang\IEEEauthorrefmark{1}, 
Hongkun Chen\IEEEauthorrefmark{2}, 
Boyan Xu\IEEEauthorrefmark{2, \Letter},
and Ruichu Cai\IEEEauthorrefmark{2},~\IEEEmembership{Senior,~IEEE}}
\IEEEauthorblockA{\IEEEauthorrefmark{1}School of Management,
Guangdong University of Technology, Guangzhou, China}
\IEEEauthorblockA{\IEEEauthorrefmark{2}School of Computer Science, Guangdong University of Technology, Guangzhou, China}
\IEEEauthorblockA{\IEEEauthorrefmark{\Letter}Corresponding author: hpakyim@gmail.com}}
\maketitle

\begin{abstract}
The advancement of medical image segmentation techniques has been propelled by the adoption of deep learning techniques, particularly UNet-based approaches, which exploit semantic information to improve the accuracy of segmentations. 
However, the order of organs in scanned images has been disregarded by current medical image segmentation approaches based on UNet. Furthermore, the inherent network structure of UNet does not provide direct capabilities for integrating temporal information. 
To efficiently integrate temporal information, we propose TP-UNet that utilizes temporal prompts, encompassing organ-construction relationships, to guide the segmentation UNet model. Specifically, our framework is featured with cross-attention and semantic alignment based on unsupervised contrastive learning to combine temporal prompts and image features effectively.
Extensive evaluations on two medical image segmentation datasets demonstrate the state-of-the-art performance of TP-UNet. Our implementation will be open-sourced after acceptance.
\end{abstract}


\section{Introduction}
\label{sec:intro}
Medical image segmentation holds a pivotal position within the realm of modern medicine, playing a fundamental role in disease diagnosis, surgical planning, and treatment monitoring~\cite{MahendraKhened2018FullyCM}. The primary objective of this task is to accurately separate and label distinct structures or tissues depicted in medical images, enabling healthcare professionals to conduct meticulous analysis and achieve precise diagnoses. Notably, with the advancements in deep learning techniques, certain networks built upon UNet and its variants have exhibited commendable segmentation accuracy by leveraging semantic information extracted from medical images~\cite{9446143}. 

\begin{figure}[!t]
\centerline{\includegraphics[width=\columnwidth]{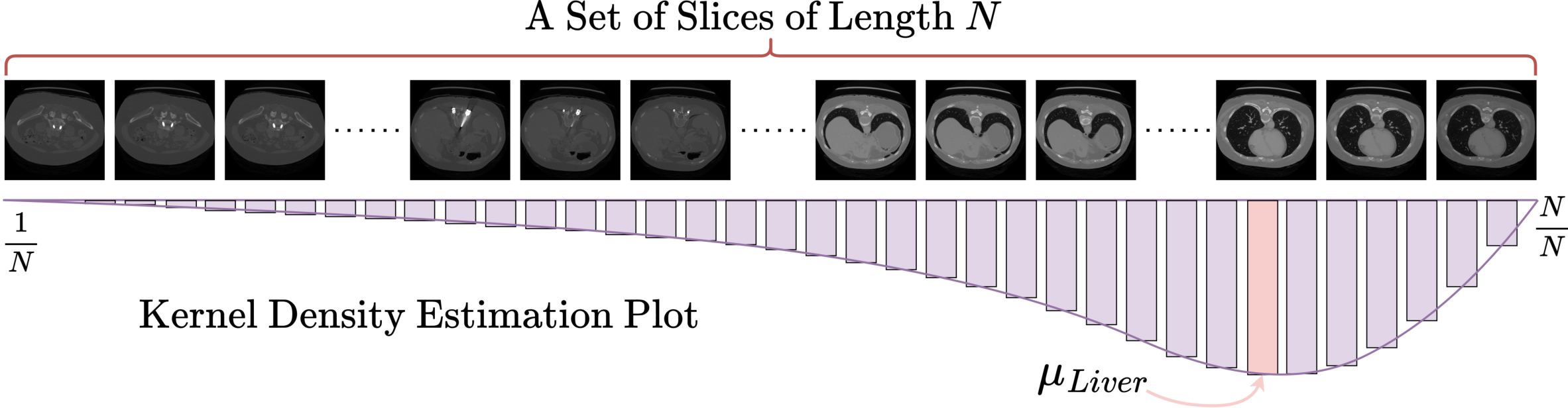}}
\caption{\textbf{The temporal information of liver.} We visualized the temporal information of the liver. From the kernel density plot of the liver occurrence probability, it can be seen that the distribution approximately follows a normal distribution \(\mathcal{N}(\mu_{Liver}, \sigma_{Liver})\) for a set of timestamps ranging from $\frac{1}{N}$ to $\frac{N}{N}$. The timestamp with the highest frequency of liver occurrence is approximately 0.78. For multiple organs, such as the three organs in the UW-Madison dataset (i.e., stomach, large intestine, and small intestine), their probabilities of occurrence at different timestamps also vary, typically $\mu_{stomach} \leq \mu_{small} \leq \mu_{large}$. This temporal information is crucial for guiding the model in segmentation tasks.}
\label{intro}
\end{figure}

Although promising results were reported, existing UNet-based approaches lack consideration for the temporal information present in scanned medical images \cite{ye2022sia}. To better understand the temporal information, we have visualized it in Fig.~\ref{intro}. Incorporating temporal information, which represents a sequence of medical images, has the potential to enhance the accuracy of medical segmentation. For instance, in a series of $N$ slices from a patient, denoted as $N_{i}^{th}/N$. Notably, certain organs such as the stomach, large intestine, and small intestine exhibit specific temporal patterns within a given interval $(N_{1}^{th}/N, N_{m}^{th}/N)$, often following a normal distribution. Specifically, their normal distributions can be expressed as \small{$\mathcal{N}_{stomach}(\mu_{stomach}, \sigma_{stomach})$}, \small{$\mathcal{N}_{large}(\mu_{large}, \sigma_{large})$}, and \small{$\mathcal{N}_{small}(\mu_{small}, \sigma_{small})$}, respectively. In imaging modalities like MRI or CT scans, where the imaging process typically progresses from top to bottom, the stomach predominantly appears in the early to mid-time intervals, while the small intestine is more prevalent in the mid-time interval, and the large intestine is primarily observed during the late-time interval. Consequently, we have $\mu_{stomach} \leq \mu_{small} \leq \mu_{large}$. Thus, within a specific temporal interval, it is essential to prioritize stomach segmentation during the early stages and focus on large intestine segmentation in the later stages to improve the performance of medical segmentation models. Despite the evident importance of temporal information in enhancing segmentation accuracy, its consideration is often overlooked in current research.\\
\begin{figure*}[!ht]
\centerline{\includegraphics[width=1.7\columnwidth]{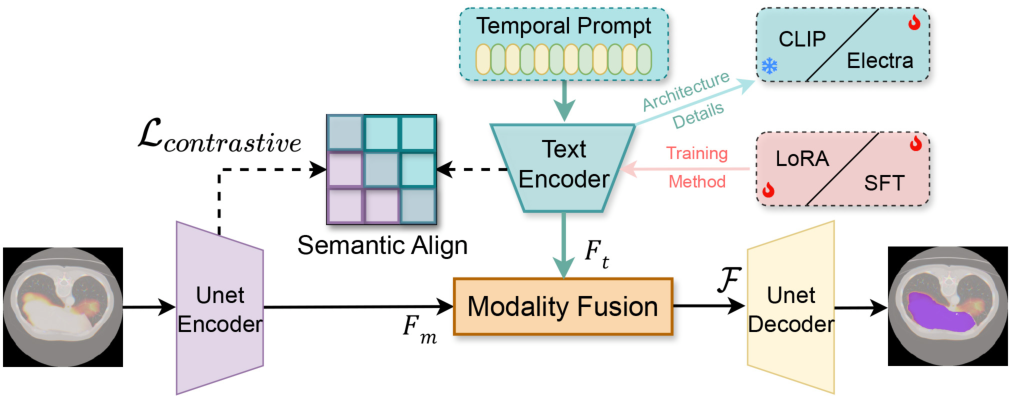}}
\caption{\textbf{The general framework of TP-UNet.} For a given medical image \(I\) that needs segmentation, TP-UNet first automatically generates its corresponding temporal prompt $P_t$. The UNet encoder then extracts features from the input medical image \(I\). These extracted features are fused with the encoded temporal prompt $F_t$. Prior to fusion, a semantic alignment operation is performed to bridge the gap between different modality encoders. Finally, the UNet decodes the fused features to produce the final masks. It should be noted that the text encoder in this study employs two architectures: CLIP and Electra. These architectures are trained using the LoRA and SFT methods, respectively.}
\label{fig1}
\end{figure*}
\indent To exploit the temporal information inherent in medical images, we propose TP-UNet, a framework that leverages temporal prompts to guide the learning process of the UNet model. The temporal prompt offers textual signals for guiding the segmentation model in learning semantic and sequential information from medical images. These textual signals first undergo high-dimensional embedding via a well-trained encoder and then interact with the feature map from the image encoder. Due to the different encoding processes of text and image, simply using a linear mapping between image embeddings and textual embeddings for interaction may lead to suboptimal fusion results or even degrade model performance~\cite{li2021align}. To address this, we perform a semantic alignment operation before the interaction between text and image modalities, utilizing unsupervised contrastive learning for text representation and image representation to narrow the semantic gap. Finally, for modality fusion, we employ a cross-attention mechanism to aggregate the aforementioned updated text representation and image representation. This process yields a unified representation that serves as input to the decoder of the UNet model. Our main contributions can be summarized as follows:
\begin{itemize}
\item We propose TP-UNet, a simple and effective framework for medical image segmentation, which can guide the segmentation model to learn the temporal information in medical images through textual prompts. 
\item We propose a two-stage process of semantic alignment and modal fusion to narrow the semantic gap between temporal prompts and image features and effectively aggregate them into a unified representation. 
\item We conducted extensive experiments on two medical image segmentation datasets, including the LITS 2017 dataset and the UW-Madison dataset. The results of the experiments demonstrated that our method achieved a new state-of-the-art (SOTA) performance.
\end{itemize}

\section{Related Work}
\label{sec:Related Work}

\subsection{Prompt Learning} 
Prompt learning is a crucial research area in the field of natural language processing (NLP). It focuses on designing effective prompts or questions to guide models in generating accurate and relevant outputs for specific tasks. By tailoring prompts to the task at hand, prompt learning helps the model concentrate on essential information and reduces the search space, thus improving model performance. This technique has proven successful in various NLP tasks, including text classification, named entity recognition, and machine translation. Recently, this approach has been applied to medical segmentation, aiming to enhance the segmentation capabilities of models through text prompts. Jie Liu et al. constructed prompts based on the names of organs that need to be segmented~\cite{liu2023clip}. Junde Wu et al., on the other hand, built prompts based on textual descriptions of organs, including their functions, shapes, and appearances, achieving significant improvements in segmentation performance~\cite{wu2023medical}. However, it is evident that neither of these approaches utilized temporal information from medical images. In this study, we designed prompts based on the temporal information of medical images, aiming to guide the segmentation model using temporal information for better performance.

\subsection{Multimodal Contrastive Learning} 
Multimodal contrastive learning is a powerful technique in the field of multimodal learning, which considers multiple modalities such as text, images, and audio. The goal of this method is to learn meaningful representations by maximizing the similarity between samples from the same category and minimizing the similarity between samples from different categories. In the context of multimodal learning, this involves aligning the representations of different modalities in a shared embedding space to capture their relationships and interactions. 
Multimodal contrastive learning leverages complementary information from multiple modalities, enhancing the model's understanding of the overall data and its representation learning capabilities. This approach has been widely applied in the medical domain. Yuhao Zhang et al. proposed ConVIRT~\cite{zhang2022contrastive}, which uses a bidirectional contrastive objective function to maximize the consistency between true matches and random pairs of images and texts, achieving unsupervised training. This method leverages paired text data across domains without requiring additional expert input. In image classification tasks, ConVIRT achieves high data efficiency, as it can achieve comparable or even better performance than models initialized with ImageNet using only 10\% labeled training data. Shih-Cheng Huang et al. proposed the GLoRIA~\cite{huang2021gloria} framework, which maximizes the correlation between medical images and texts using global and local contrastive loss functions, leading to improved performance in downstream tasks. In summary, multimodal contrastive learning can enhance data efficiency, align semantic information between different modalities, and improve performance in downstream tasks. The contrastive learning used in this paper addresses the issue of similarity between different modalities, resulting in better generalization and segmentation performance for downstream medical image segmentation tasks.

\section{Methodology}
\label{sec:Methodology}
In this section, we introduce the TP-UNet model (as shown in Fig~\ref{fig1}), which addresses the issue of temporal information forgetting in medical image analysis by designing the Temporal Prompt module. Additionally, we utilize the Semantic Align module to bridge the semantic gap between temporal prompt and image modalities. The combined effect of these two key components significantly enhances the performance of TP-UNet in medical image segmentation, enabling more precise and consistent segmentation of dynamic images.
\subsection{Temporal Prompt}
Temporal information plays a crucial role in improving model segmentation performance. we devised a set of prompts to guide the model in understanding the temporal information of medical images. In this study, temporal information is represented as $N^{th}_i/N$, indicating that this information is mapped to the interval [0, 1]. The occurrence probability of organs follows a normal distribution within this interval, allowing the model to comprehend the varying probabilities of organ appearance at different timestamps, thereby adjusting its focus on different organs accordingly. The temporal prompt template defined in this study is as follows: \textit{"This is \{an MRI / a CT\} of the \{organ\} with a segmentation period of \{$N^{th}_i/N$\}."} Here, the type of medical image and the organ can be selected, while $N$ is determined by the size of a set of slices. In this study, the temporal prompts are automatically generated. Before being input into TP-UNet, a set of prompts is automatically created using the numpy and pandas libraries based on the type of image selected by the physician. In specific situations, radiologists can also choose the range of timestamps for segmentation by dragging to select the desired range. This allows TP-UNet to save a significant amount of time during inference by generating prompts only for the specified range of slices. The generation of a temporal prompt for a single set of slices takes less than 1ms, which is highly significant for clinical applications in radiology.
\subsection{Multimodal Encoder}
We first define the input medical image as \(I\) and the generated temporal prompt as \(P_t\). For the text-based temporal prompt and the medical image requiring segmentation, we designed a multimodal encoder. For the input text modality \(P_t\), we adopted two encoding methods. The first method utilizes the popular multimodal text encoder CLIP~\cite{radford2021learning}. While CLIP performs well with general natural language, directly applying it to medical text may introduce a domain gap. Therefore, we employed parameter-efficient fine-tuning (PEFT)~\cite{peft} using the LoRA~\cite{hu2021lora} method to adapt CLIP more effectively to our task. The second text encoder we used is Electra~\cite{clark2020electra}, another popular text encoder. We compared the performance of these two encoders in the experiments section. We performed supervised fine-tuning (SFT)~\cite{ouyang2022training} on the pre-trained Electra model. Both fine-tuned text encoders demonstrated good performance.

For the medical image modality, we used the conventional UNet method for segmentation. We integrated the low-level semantics extracted by UNet with the temporal prompt to guide the model for more effective segmentation based on temporal information. The details of the fusion method will be presented in Section~\ref{Modality Fusion}.

\subsection{Semantic Align}
In this context, we define the encoded image feature $F_m \in \mathbb{R}^{B \times C \times H \times W}$ and the textual temporal prompt encoded feature $F_t \in \mathbb{R}^{B \times L \times D}$. Before modality fusion, $I$ undergoes an UNet encoder block, while $P_t$ via the text encoder. Due to the disparate network architectures of the two models, they originate from different semantic spaces, potentially leading to a performance decrease after fusion. Therefore, it becomes essential to align the semantics of $F_m$ and $F_t$ before modality fusion. To achieve this, we introduce a semantic align module, aiming to bring semantically similar pairs of $F_{mi}$ and $F_{ti}$ closer together in a batch, while pushing semantically dissimilar $F_{mi}$ and non-corresponding $F_{tj}$ further apart. Consequently, the first contrastive loss function is an image-to-text contrastive loss for the i-th pair:

\begin{equation}
\scalebox{1.5}{$
\ell_{i}^{(F_m \rightarrow F_t)}=-\log \frac{\exp \left(\left\langle F_{mi}, F_{ti}\right\rangle / \tau\right)}{\sum_{k=1}^{N} \exp \left(\left\langle F_{mi}, F_{tk}\right\rangle / \tau\right)}
$}
\label{num1}
\end{equation}

\begin{equation}
\scalebox{1.2}{$
\begin{aligned}
\begin{array}{l}
\langle F_{mi}, F_{ti}\rangle= F_{mi}^{\top} F_{ti} /\|F_{mi}\|\|F_{ti}\|
\end{array}
\end{aligned}
$}
\label{num2}
\end{equation}
where $\tau \in \mathbb{R}^{+}$ represents a temperature parameter.\\
The second loss function is a text-to-image contrastive loss for the i-th pair:
\begin{equation}
\scalebox{1.4}{$
\begin{aligned}
\begin{array}{l}
\ell_{i}^{(F_t \rightarrow F_m)}=-\log \frac{\exp \left(\left\langle F_{ti}, F_{mk}\right\rangle / \tau\right)}{\sum_{k=1}^{N} \exp \left(\left\langle F_{ti}, F_{mk}\right\rangle / \tau\right)} 
\end{array}
\end{aligned}
$}
\label{num3}
\end{equation}
In the end, the loss we need to optimize is:
\begin{equation}
\begin{aligned}
\begin{array}{l}
\mathcal{L}_{contrastive}=\frac{1}{N} \sum_{i=1}^{N}\left(\lambda \ell_{i}^{(F_m \rightarrow F_t)}+(1-\lambda) \ell_{i}^{(F_t \rightarrow F_m)}\right) 
\end{array}
\label{num4}
\end{aligned}
\end{equation}
where $\lambda \in [0,1]$ is a scalar weight. Through the semantic alignment module, the semantic representations of different modalities are aligned. This lays a solid foundation for the subsequent modality fusion.

\subsection{Modality Fusion}
\label{Modality Fusion}
Temporal prompt is crucial to improve the segmentation performance of the model. Consequently, there should be increased emphasis on the design of modality fusion between the temporal prompt modality and the visual modality. Therefore, we designed the cross-attention mechanism, which can be represented as follows:
\begin{equation}
\begin{aligned}
\begin{array}{l}
\mathcal{F}=\operatorname{softmax}\left(\frac{\left([F_m';F_t']\mathbf{W}^{Q}\left([F_m';F_t'] \mathbf{W}^{K}\right)^{\top}\right)}{\sqrt{d_{k}}}\right) [F_m';F_t'] \mathbf{W}^{V}
\end{array}
\label{num5}
\end{aligned}
\end{equation}
Where $[;]$ represents the concatenate operation, $\mathcal{F}$ is the pixel-wise attention map, and $d_k$ is a scaling factor. $F_m'$ and $F_t'$ are the projections of $F_m$ and $F_t$. $\mathbf{W}^{Q}$, $\mathbf{W}^{K}$ and $\mathbf{W}^{V}$ are the corresponding weight matrices.
Finally, the feature map $\mathcal{F}$ is concatenated with the first-level skip-connection feature map of the UNet. It undergoes a convolutional layer and a ReLU activation function before passing through a 1$\times$1 convolutional layer to generate the final segmentation image.

\begin{table*}[!ht]
\begin{center}
\caption{Main Result on UW-Madison Dataset}
\label{table1}
\vspace{-3mm}
\setlength{\tabcolsep}{1.7mm}{
\resizebox{\textwidth}{!}{
\renewcommand\arraystretch{1.2}
\begin{tabular}{c|c|c|c|c|c|c|c|c|c}
\hline
\hline
\multicolumn{1}{c|}{\multirow{2}*{Model}} & \multirow{2}*{Backbone} & \multicolumn{2}{c|}{Large Intestine} & \multicolumn{2}{c|}{Small Intestine} & \multicolumn{2}{c|}{Stomach} & \multicolumn{2}{c}{Average} \\
\cline{3-10}
{}&{} & Dice~$\uparrow$ & Jacc~$\uparrow$ & Dice~$\uparrow$ & Jacc~$\uparrow$ & Dice~$\uparrow$ & Jacc~$\uparrow$ & Dice~$\uparrow$ & Jacc~$\uparrow$\\
\hline \hline
\small UNet~\cite{ronneberger2015u} & \small VGG16~\cite{simonyan2014very} & \multicolumn{1}{c}{\small 0.8709} & \multicolumn{1}{c|}{\small 0.8329} & \multicolumn{1}{c}{\small 0.8546} & \multicolumn{1}{c|}{\small 0.8187} & \multicolumn{1}{c}{\small 0.9210} & \multicolumn{1}{c|}{\small 0.9001} & \multicolumn{1}{c}{\small 0.8822} & \multicolumn{1}{c}{\small 0.8506}\\
\small UNet~\cite{ronneberger2015u} & \small Resnet50~\cite{he2016deep} & \multicolumn{1}{c}{\small 0.8621} & \multicolumn{1}{c|}{\small 0.8211} & \multicolumn{1}{c}{\small 0.8487} & \multicolumn{1}{c|}{\small 0.8122} & \multicolumn{1}{c}{\small 0.9106} & \multicolumn{1}{c|}{\small 0.8880} & \multicolumn{1}{c}{\small 0.8738} & \multicolumn{1}{c}{\small 0.8405}\\
\small UNet++~\cite{zhou2019UNet++} & \small VGG16~\cite{simonyan2014very} &\multicolumn{1}{c}{\small  0.8825} & \multicolumn{1}{c|}{\small 0.8463} & \multicolumn{1}{c}{\small 0.8738} & \multicolumn{1}{c|}{\small 0.8394} & \multicolumn{1}{c}{\small 0.9394} & \multicolumn{1}{c|}{\small 0.9201} & \multicolumn{1}{c}{\small 0.8986} & \multicolumn{1}{c}{\small 0.8686}\\
\small UNet++~\cite{zhou2019UNet++} & \small Resnet50~\cite{he2016deep} & \multicolumn{1}{c}{\small 0.8613} & \multicolumn{1}{c|}{\small 0.8223} & \multicolumn{1}{c}{\small 0.8537} & \multicolumn{1}{c|}{\small 0.8172} & \multicolumn{1}{c}{\small 0.9178} & \multicolumn{1}{c|}{\small 0.8970} & \multicolumn{1}{c}{\small 0.8776} & \multicolumn{1}{c}{\small 0.8455}\\
\small Attention UNet~\cite{oktay2018attention} & \small VGG16~\cite{simonyan2014very} & \multicolumn{1}{c}{\small 0.8925} & \multicolumn{1}{c|}{\small 0.8580} & \multicolumn{1}{c}{\small 0.8857} & \multicolumn{1}{c|}{\small 0.8519} & \multicolumn{1}{c}{\small 0.9472} & \multicolumn{1}{c|}{\small 0.9292} & \multicolumn{1}{c}{\small 0.9085} & \multicolumn{1}{c}{\small 0.8797}\\
\small scSE UNet~\cite{roy2018concurrent} & \small VGG16~\cite{simonyan2014very} & \multicolumn{1}{c}{\small 0.8909} & \multicolumn{1}{c|}{\small 0.8561} & \multicolumn{1}{c}{\small 0.8750} & \multicolumn{1}{c|}{\small 0.8398} & \multicolumn{1}{c}{\small 0.9491} & \multicolumn{1}{c|}{\small 0.9312} & \multicolumn{1}{c}{\small 0.9050} & \multicolumn{1}{c}{\small 0.8757}\\
\small Trans UNet~\cite{chen2021transUNet} & \small Vision Transformer~\cite{dosovitskiy2020image} & \multicolumn{1}{c}{\small 0.8983} & \multicolumn{1}{c|}{\small 0.8597} & \multicolumn{1}{c}{\small 0.8847} & \multicolumn{1}{c|}{\small 0.8543} & \multicolumn{1}{c}{\small 0.9356} & \multicolumn{1}{c|}{\small 0.9185} & \multicolumn{1}{c}{\small 0.9062} & \multicolumn{1}{c}{\small 0.8775}\\
\small Swin UNet~\cite{cao2023swin} & \small Swin Transformer~\cite{liu2021swin} & \multicolumn{1}{c}{\small 0.9034} & \multicolumn{1}{c|}{\small 0.8704} & \multicolumn{1}{c}{\small 0.8908} & \multicolumn{1}{c|}{\small 0.8617} & \multicolumn{1}{c}{\small 0.9467} & \multicolumn{1}{c|}{\small 0.9279} & \multicolumn{1}{c}{\small 0.9136} & \multicolumn{1}{c}{\small 0.8819}\\
\hline
\small TP-UNet\ $_{Electra}$ & \small VGG16~\cite{simonyan2014very} & \multicolumn{1}{c}{\small 0.9170} & \multicolumn{1}{c|}{\small 0.8781} & \multicolumn{1}{c}{\small 0.9078} & \multicolumn{1}{c|}{\small 0.8704} & \multicolumn{1}{c}{\small 0.9551} & \multicolumn{1}{c|}{\small 0.9343} & \multicolumn{1}{c}{\small 0.9266} &\multicolumn{1}{c}{\small 0.8943}\\ 
\small TP-UNet\ $_{CLIP}$ & \small VGG16~\cite{simonyan2014very} & \multicolumn{1}{c}{\small $\boldsymbol{0.9190}$ } & \multicolumn{1}{c|}{\small $\boldsymbol{0.8810}$ }& \multicolumn{1}{c}{\small $\boldsymbol{0.9102}$ }& \multicolumn{1}{c|}{\small $\boldsymbol{0.8727}$} & \multicolumn{1}{c}{\small $\boldsymbol{0.9566}$} & \multicolumn{1}{c|}{\small $\boldsymbol{0.9358}$ }& \multicolumn{1}{c}{\small $\boldsymbol{0.9286}$} &\multicolumn{1}{c}{\small $\boldsymbol{0.8965}$}\\
\hline \hline
\end{tabular}}}
\end{center}
\end{table*}

\begin{table}[h]
\caption{Main Result on LITS Dataset}
\label{Main LITS}
\vspace{-3mm}
\begin{center}
\setlength{\tabcolsep}{0.5mm}{
\renewcommand\arraystretch{1.2}
\begin{tabular}{c|c|c|c}
\hline
\hline
\multicolumn{1}{c|}{\multirow{2}*{Model} } & \multirow{2}*{BackBone} & \multicolumn{2}{c}{Liver}\\
\cline{3-4}
{}&{}&Dice~$\uparrow$ & Jacc~$\uparrow$\\
\hline
\hline
\small UNet~\cite{ronneberger2015u} & \small VGG16~\cite{simonyan2014very} & \multicolumn{1}{c}{\small 0.8517} & \small 0.8147\\
\small UNet~\cite{ronneberger2015u} & \small Resnet50~\cite{he2016deep} &\multicolumn{1}{c}{\small 0.7223} &\small 0.6848\\
\small UNet++~\cite{zhou2019UNet++} & \small VGG16~\cite{simonyan2014very} & \multicolumn{1}{c}{\small 0.8486} & \small 0.8098\\
\small UNet++~\cite{zhou2019UNet++} & \small Resnet50~\cite{he2016deep} &\multicolumn{1}{c}{\small  0.8135} &\small  0.7751\\
\small Attention UNet~\cite{oktay2018attention} & \small VGG16~\cite{simonyan2014very} &\multicolumn{1}{c}{\small  0.8500} &\small 0.8137\\
\small scSE UNet~\cite{roy2018concurrent} & \small VGG16~\cite{simonyan2014very} &\multicolumn{1}{c}{\small  0.8498} & \small 0.8143\\
\small Trans UNet~\cite{chen2021transUNet} & \small Vision Transformer~\cite{dosovitskiy2020image} &\multicolumn{1}{c}{\small  0.8069} &\small  0.7699\\
\small Swin UNet~\cite{cao2023swin} & \small Swin Transformer~\cite{liu2021swin} &\multicolumn{1}{c}{\small  0.8204} &\small  0.7833\\
\hline
\small TP-UNet~$_{Electra}$  & \small VGG16~\cite{simonyan2014very} &\multicolumn{1}{c}{\small  $\boldsymbol{0.9125}$ }&\small  $\boldsymbol{0.8780}$\\
\small TP-UNet~$_{CLIP}$  & \small VGG16~\cite{simonyan2014very} &\multicolumn{1}{c}{\small  0.8657 }&\small 0.8269\\
\hline
\hline
\end{tabular}}
\end{center}
\end{table}

\section{Experiments}
\label{sec:Experiments}
\subsection{Experimental Settings}
\subsubsection{Dataset}

\textbf{UW-Madison Dataset Collection:} \footnote[1]{\href{https://www.kaggle.com/competitions/uw-madison-gi-tract-image-segmentation/data}{\textcolor{pink}{https://www.kaggle.com/datasets/uwmgi-mask-dataset}}} The dataset originates from MRI scan images of multiple patients at the Carbone Cancer Center of the University of Wisconsin-Madison. It primarily consists of MRI images of the colon and stomach regions. We call it the UW-Madison dataset~\cite{uw-madison-gi-tract-image-segmentation} and divide it according to a 7: 1: 2 ratio into training, validation, and testing set. Among them, the training set contains 26746 images, the validation set contains 3820 images, and the testing set contains 7642 images. \\
\textbf{LITS Dataset Collection:} \footnote[2]{\href{https://competitions.codalab.org/competitions/17094}{\textcolor{pink}{https://competitions.codalab.org/competitions/LITS 2017}}}
LITS~\cite{bilic2023liver} is an acronym for Liver Tumor Segmentation Benchmark. The data and segmentations are provided by various clinical sites around the world. The dataset contains CT scans of 130 patients. But these scans are 3D nii files, what we need are 2D slices. We divide the LITS dataset into 58638 2D slices, but a large number of 2D slices are also redundant. We finally selected 10967 2D slices with sequence information, and divided these slices according to a 7: 1: 2 ratio into training, validation, and testing set.
\subsubsection{Implementation details}
We choose the Adaptive Momentum Estimation  with a weight decay of 0.000001 as the training optimizer. Meanwhile, the initial learning rate is 0.00003, and the weights change with the cosine annealing learning rate; the initial temperature is 25, and the maximum temperature is 96.875. We use the PyTorch training framework and some data augmentation methods, such as CoarseDropout, HorizontalFlip, and ShiftScaleRotate. The loss function uniformly adopts the average value of Binary Cross-Entropy ($\mathcal{L}_{BCE}$) and Tversky loss ($\mathcal{L}_{Tversky}$), $\mathcal{L}_{Tversky}$ is the loss function to solve imbalanced classification problems. 
\subsubsection{Evaluation metrics}
We use the Jaccard coefficient and the Dice coefficient to evaluate the performance of the model, which can measure the performance of the model by calculating the similarity between the ground-truth annotation and the predicted annotation. Their calculation can be expressed as follows:

\begin{equation}
\begin{aligned}
\begin{array}{l}
Jaccard =\frac{A \cap B}{A \cup B} \vspace{1ex},\\
Dice = \frac{2(A \cap B)}{A+B},
\end{array}
\label{num6}
\end{aligned}
\end{equation}
where $A$ and $B$ are binary matrices representing the ground-truth annotation and the predicted annotation, respectively.


\subsection{Comparison with the Baselines}
To demonstrate the effectiveness of TP-UNet, we conducted extensive experiments on two different datasets. Several commonly used medical image segmentation models are selected for experimental comparison. We use the Jaccard coefficient and the Dice coefficient to evaluate the performance.

As shown in Table~\ref{table1}, TP-UNet achieved the best performance in all three organ categories (large intestine, small intestine, and stomach) as well as the overall average performance on the UW-Madison dataset. Compared to UNet, the Dice score improved by an average of 4.44\%, with the most significant improvement of 5.32\% in the Small Intestine category. In addition to UNet, we also compared our TP-UNet with several other methods listed in Table \ref{table1}. TP-UNet outperformed the current state-of-the-art (Swin UNet~\cite{cao2023swin}) by 1.3\% in the Dice score, with the most significant improvement of 1.7\% in the Small Intestine category.\\
\indent We also conducted experiments on the LITS 2017 dataset. As shown in Table~\ref{Main LITS}, our method achieved the highest Dice and Jaccard scores for liver segmentation. Compared to UNet, the Dice score for the liver increased by 6.08\%, and the Jaccard score increased by 6.33\%. On the LITS 2017 dataset, TP-UNet outperformed the current state-of-the-art method by 9.21\% in Dice score, with the most significant improvement of 9.47\% in the Small Intestine category.



\begin{table*}[h]

\begin{center}
\caption{ablation study}
\label{table3}
\vspace{-3mm}
\setlength{\tabcolsep}{1.7mm}{
\resizebox{\textwidth}{!}{
\renewcommand\arraystretch{1.2}
\begin{tabular}{l|c|c|c|c|c|c|c|c|c|c}
\hline
\multicolumn{1}{c|}{}&\multicolumn{8}{c|}{UW-Madison Dataset}&\multicolumn{2}{c}{LITS 2017}\\
\hline
\hline
\multirow{2}*{Model} & \multicolumn{2}{c|}{Large Intestine} & \multicolumn{2}{c|}{Small Intestine} & \multicolumn{2}{c|}{Stomach} & \multicolumn{2}{c|}{Average} & \multicolumn{2}{c}{Liver}\\
\cline{2-11}
{} & Dice~$\downarrow$ & Jacc~$\downarrow$ & Dice~$\downarrow$ & Jacc~$\downarrow$& Dice~$\downarrow$ & Jacc~$\downarrow$& Dice~$\downarrow$ & Jacc~$\downarrow$& Dice~$\downarrow$ & Jacc~$\downarrow$\\
\hline
\midrule
\textbf{TP-UNet~$_{Electra}$} &\multicolumn{1}{c}{\small 0.9170} & \multicolumn{1}{c}{\small 0.8781} & \multicolumn{1}{c}{\small 0.9078} & \multicolumn{1}{c}{\small 0.8704} & \multicolumn{1}{c}{\small 0.9551} & \multicolumn{1}{c}{\small 0.9343} & \multicolumn{1}{c}{\small 0.9266} & \small 0.8943& \multicolumn{1}{c}{\small 0.9125} & \multicolumn{1}{c}{\small 0.8780}\\ \midrule
~~~~ w/o $\text{Temporal Information}$ &\multicolumn{1}{c}{\small 0.8924} & \multicolumn{1}{c}{\small 0.8545} & \multicolumn{1}{c}{\small 0.8787} & \multicolumn{1}{c}{\small 0.8423} & \multicolumn{1}{c}{\small 0.9306} & \multicolumn{1}{c}{\small 0.9117} & \multicolumn{1}{c}{\small 0.9048} & \small 0.8733& \multicolumn{1}{c}{\small 0.8735} & \multicolumn{1}{c}{\small 0.8396}\\

~~~~ w/o Temporal Prompt &\multicolumn{1}{c}{\small 0.8797} & \multicolumn{1}{c}{\small 0.8427} & \multicolumn{1}{c}{\small 0.8666} & \multicolumn{1}{c}{\small 0.8305} & \multicolumn{1}{c}{\small 0.9306} & \multicolumn{1}{c}{\small 0.9117} & \multicolumn{1}{c}{\small 0.8923} & \small 0.8617& \multicolumn{1}{c}{\small 0.8589} & \multicolumn{1}{c}{\small 0.8246}\\

~~~~ w/o Semantic Align &\multicolumn{1}{c}{\small 0.9082} & \multicolumn{1}{c}{\small 0.8683} & \multicolumn{1}{c}{\small 0.8958} & \multicolumn{1}{c}{\small 0.8586} & \multicolumn{1}{c}{\small 0.9455} & \multicolumn{1}{c}{\small 0.9227} & \multicolumn{1}{c}{\small 0.9165} & \small 0.8832& \multicolumn{1}{c}{\small 0.9053} & \multicolumn{1}{c}{\small 0.8681}\\

~~~~ w/o Modality Fusion &\multicolumn{1}{c}{\small 0.9101} & \multicolumn{1}{c}{\small 0.8684} & \multicolumn{1}{c}{\small 0.8996} & \multicolumn{1}{c}{\small 0.8623} & \multicolumn{1}{c}{\small 0.9434} & \multicolumn{1}{c}{\small 0.9237} & \multicolumn{1}{c}{\small 0.9177} & \small 0.8848& \multicolumn{1}{c}{\small 0.9047} & \multicolumn{1}{c}{\small 0.8709}\\

\bottomrule
\end{tabular}}}
\end{center}
\end{table*}

\begin{figure*}[!ht]
\centerline{\includegraphics[width=1.38\columnwidth]{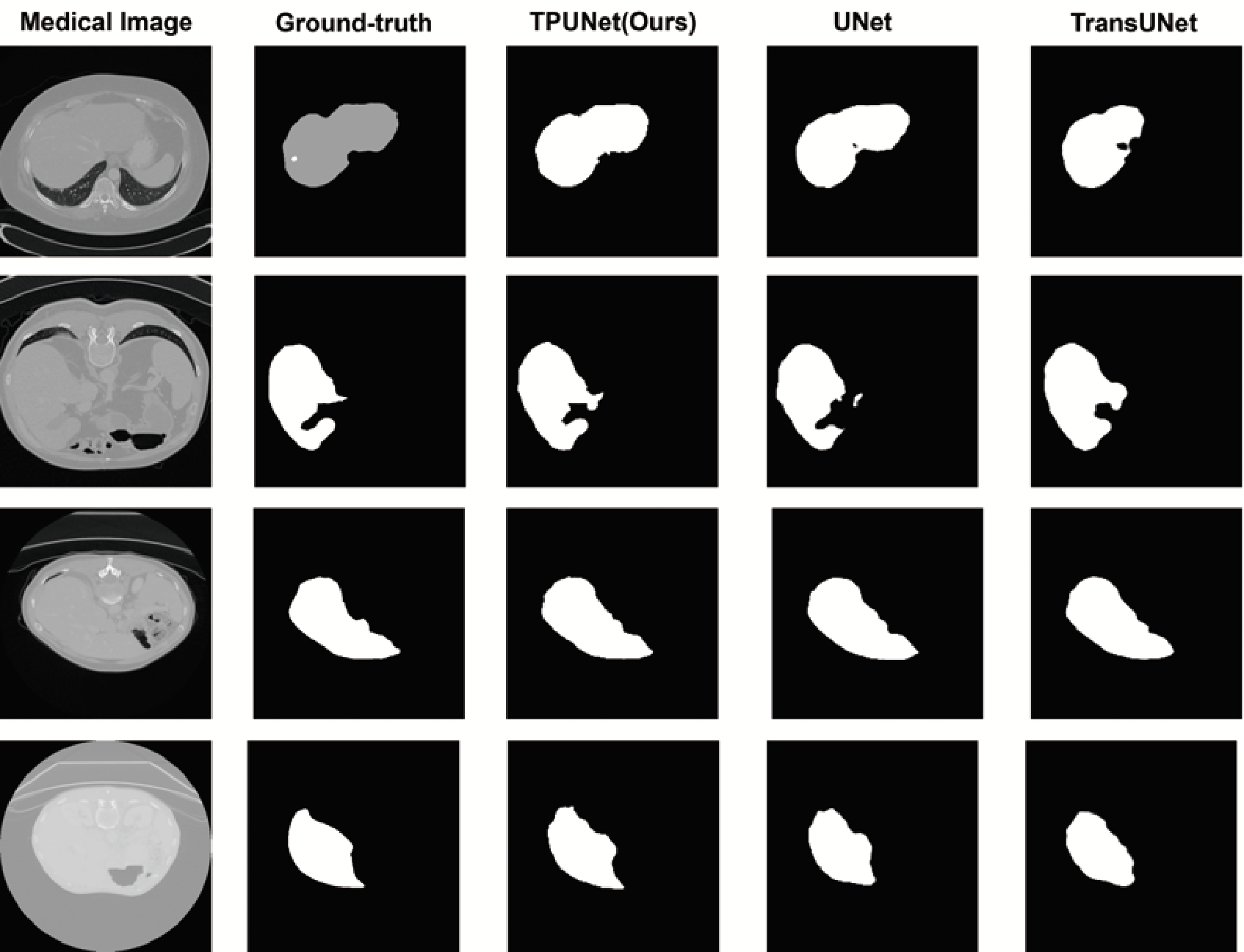}}
\caption{\textbf{Case Study.} We conducted four case studies on the LITS dataset. From the results of the qualitative analysis, our method achieved excellent performance.}
\label{fig2}
\end{figure*}

\subsection{Case Study}
This case study provides a comparative analysis of TP-UNet and other baseline methods, as shown on Figure \ref{fig2}. TP-UNet, our proposed model, demonstrated superior performance over the traditional U-Net and the transformer-based TransUNet in the segmentation of CT scans.

We visualized three segmentation results on LITS2017 dataset, including those from our method. From a visual standpoint, our results closely resemble the ground truth. Our algorithm demonstrates superior performance in areas where many commonly used segmentation techniques fail, especially when dealing with details that are imperceptible to the naked eye.

\subsection{Ablation Study}
We conducted ablation experiments on the UW-Madison dataset and the LITS2017 dataset to demonstrate the effectiveness of the TP-UNet methods.

The experimental results are shown in Table~\ref{table3}, where a lower score indicates a greater contribution of the module to the TP-UNet model. First, we verified the effectiveness of the temporal information by removing the timestamp from the temporal prompt, fixing the prompt template to \textit{"This is {an MRI / a CT} of the {organ}"}. Keeping other settings unchanged, we observed a 2.1\% decrease in the mDice score on the UW-Madison dataset. This further demonstrates the effectiveness of incorporating temporal information, which significantly helps guide the model in enhancing segmentation performance.

Next, we removed the entire temporal prompt and did not use a text encoder. Consequently, the modality fusion changed to a self-attention mechanism. From the results, we observed a significant decrease of 5.36\% in the mDice score on the LITS dataset. This demonstrates that the temporal prompt not only provides valuable temporal information but also that the selected organ and image type are beneficial for segmentation.

We also explored the effectiveness of the semantic alignment module. We performed modality fusion directly without semantic alignment beforehand. The results showed that the mDice score on the UW-Madison dataset decreased by 1.01\%. This demonstrates that semantic alignment is essential for multimodal fusion, as it helps reduce the domain gap between different modality encoders. This improvement enhances the efficiency of multimodal fusion and the overall performance of the model.

We also investigated the performance of the modality fusion module. We replaced the modality fusion with a direct concatenation of the inputs before the final segmentation decoder. The results indicated that the modality fusion module is indispensable for the TP-UNet framework. The proposed modality fusion method in this paper demonstrates both efficiency and superior performance.

Through the above four sets of experiments, we can identify the effectiveness of the Temporal Prompt, the necessity of the semantic alignment module, and the efficiency of the modality fusion module. These three components are indispensable for TP-UNet and are crucial for its superior performance.


\section{Conclusion}
\label{sec:Conclusion}
In this paper, we propose a temporal prompt guiding framework for medical image segmentation, which guides the learning of a segmentation model to the inherent temporal information of scanned images through straightforward temporal prompts. In addition, we further propose a two-stage process including semantic alignment and modality fusion to aggregate temporal prompts textual representation and image representations via multimodal contrast learning and cross-attention mechanisms.
Our proposed framework also validates the necessity of temporal information in medical image segmentation tasks through promising results. In future work, we will extend the proposed framework to more complex scenarios.

\newpage
\bibliography{conference_101719.bib}
\bibliographystyle{IEEEtran.bst}

\end{document}